\documentclass[10pt,twocolumn,letterpaper]{article}

\usepackage{cvpr}
\usepackage{times}
\usepackage{epsfig}
\usepackage{graphicx}
\usepackage{amsmath}
\usepackage{amssymb}
\usepackage{algorithm}
\usepackage{algorithmic}
\usepackage{bm, multirow}
\usepackage{enumitem}

\usepackage{color,soul}

\usepackage{tabularx} % for tables
\usepackage{rotating}
\newcolumntype{L}[1]{>{\hsize=#1\hsize\raggedright\arraybackslash}X}%
\newcolumntype{R}[1]{>{\hsize=#1\hsize\raggedleft\arraybackslash}X}%
\newcolumntype{C}[1]{>{\hsize=#1\hsize\centering\arraybackslash}X}%

\newcommand{\bK}{\mathbf{K}}
\newcommand{\bx}{\bm{x}}
\newcommand{\by}{\bm{y}}
\newcommand{\bz}{\bm{z}}
\newtheorem{tm}{Theorem}
\DeclareMathOperator*{\argmin}{arg\,min}
\graphicspath{{figs/}}

\usepackage[pagebackref=true,breaklinks=true,letterpaper=true,colorlinks,bookmarks=false]{hyperref}

\cvprfinalcopy % *** Uncomment this line for the final submission

\def\cvprPaperID{6840} % *** Enter the CVPR Paper ID here

\ifcvprfinal\fi

\begin{document}

%%%%%%%%% TITLE
\title{End-to-End Learnable Geometric Vision by Backpropagating PnP Optimization}

\author{Bo Chen$^1$ \quad \'{A}lvaro Parra$^1$ \quad  Jiewei Cao$^1$ \quad Nan Li$^2$ \quad Tat-Jun Chin$^1$\\
$^1$The University of Adelaide \quad $^2$Shenzhen University\\
{\tt\small \{bo.chen, alvaro.parrabustos,  jiewei.cao, tat-jun.chin\}@adelaide.edu.au \quad nan.li@szu.edu.cn}
}

\maketitle
%\thispagestyle{empty}

%%%%%%%%% ABSTRACT
\begin{abstract}
Deep networks excel in learning patterns from large amounts of data. On the other hand, many geometric vision tasks are specified as optimization problems. To seamlessly combine deep learning and geometric vision, it is vital to perform learning and geometric optimization end-to-end. Towards this aim, we present BPnP, a novel network module that backpropagates gradients through a Perspective-n-Points (PnP) solver to guide parameter updates of a neural network. Based on implicit differentiation, we show that the gradients of a ``self-contained" PnP solver can be derived accurately and efficiently, as if the optimizer block were a differentiable function. We validate BPnP by incorporating it in a deep model that can learn camera intrinsics, camera extrinsics (poses) and 3D structure from training datasets. Further, we develop an end-to-end trainable pipeline for object pose estimation, which achieves greater accuracy by combining feature-based heatmap losses with 2D-3D reprojection errors. Since our approach can be extended to other optimization problems, our work helps to pave the way to perform learnable geometric vision in a principled manner. Our PyTorch implementation of BPnP is available on \url{http://github.com/BoChenYS/BPnP}.
\end{abstract}

%%%%%%%%% BODY TEXT
\section{Introduction}

The success of deep learning is due in large part to its ability to learn patterns from vast amounts of training data. Applications that have benefited from this ability include object detection and image segmentation~\cite{krizhevsky12,he17}. Fundamentally, such problems can often be formulated as classification/regression problems, which facilitates suitable objective functions for backpropagation learning~\cite{lecun15}.

On the other hand, there are many important computer vision tasks that are traditionally formulated as geometric optimization problems, e.g., camera localization/pose estimation, 3D reconstruction, point set registration. A common property in these optimization problems is the minimization of a residual function (e.g., sum of squared reprojection errors) defined over geometric quantities (e.g., 6DOF camera poses), which are not immediately amenable to backpropagation learning. This limits the potential of geometric vision tasks to leverage large datasets.

A straightforward solution towards ``learnable" geometric vision is to replace the ``front end" modules (e.g., image feature detection and matching) using a deep learning alternative~\cite{Thewlis2016fully,Yi2016lift,Suwajanakorn2018discovery}. However, this does not allow the ``back end" steps (e.g., searching for optimal geometric quantities) to influence the training of the neural network parameters.

On the other extreme, end-to-end methods have been devised~\cite{kendall2015posenet,Kendall2016modelling,kendall2017geometric,brahmbhatt2018geometry,naseer2017deep,walch2017image,wu2017delving,Cai2018hybrid} that bypass geometric optimization, by using fully connected layers to compute the geometric quantity (e.g., 6DOF camera pose) from a feature map derived from previous layers. However, it has been observed that these methods are equivalent to performing image retrieval~\cite{Sattler2019understanding}, which raises questions on their ability to generalize. Also, such end-to-end methods do not explicitly exploit established methods from geometric vision~\cite{hartley2003multiple}, such as solvers for various well-defined tasks.

To benefit from the strengths of deep learning \emph{and} geometry, it is vital to combine them in a mutually reinforcing manner. One approach is to incorporate a geometric optimization solver in a deep learning architecture, and allow the geometric solver to participate in guiding the updates of the neural network parameters, thereby realising end-to-end learnable geometric vision. The key question is how to compute gradients from a ``self-contained" optimizer.

A recent work towards the above goal is differentiable RANSAC~\cite{Brachmann2017dsac, brachmann2019expert, brachmann2019neural}, which was targeting at the camera localization task. A perspective-n-point (PnP) module was incorporated in a deep architecture, and the derivatives of the PnP solver are calculated using central differences~\cite{richardson1954introduction} to enable parameter updates in the rest of the pipeline. However, such an approach to compute gradients is inexact and time consuming because, in order to obtain each partial derivative, it requires solving PnP at values that lie to the left and right of the input.

Other approaches to derive gradients from an independent optimization block for backpropagation learning~\cite{gould2016differentiating,amos2017optnet} conduct implicit differentiation~\cite[Chap.~8]{binmore83}. Briefly, in the context of end-to-end learning, the gradient of the optimization routine with respect to the input variables can be computed via partial derivatives of the stationary constraints of the optimization problem (more details in Sec.~\ref{sec:bpnp}). The gradient can then be backpropagated to the previous layers for parameter updates. A number of motivating examples and applications were explored in~\cite{gould2016differentiating,amos2017optnet}. However, larger-scale experiments in the context of specific geometric vision problems, and benchmarking against other end-to-end learning alternatives, were unavailable in~\cite{gould2016differentiating,amos2017optnet}. It is worth noting that implicit differentiation of optimization subroutines has been explored previously in several computer vision applications~\cite{tappen07,eriksson10,schmidt14} (also earlier in~\cite[Chap.~5]{faugeras93}).

\paragraph{Contributions}

Our main contribution is a novel network module called \emph{BPnP} that incorporates a PnP solver. BPnP backpropagates the gradients through the PnP ``layer" to guide the updates of the neural network weights, thereby achieving end-to-end learning using an established objective function (sum of squared 2D-3D reprojection errors) and solver from a geometric vision problem. Despite incorporating only a PnP solver, we show how BPnP can be used to learn effective deep feature representations for multiple geometric vision tasks (pose estimation, structure-from-motion, camera calibration). We also compare our method against state-of-the-art methods for geometric vision tasks. Fundamentally, our method is based on implicit differentiation; thus our work can be seen as an application of~\cite{gould2016differentiating,amos2017optnet} to geometric vision learning.

%------------------------------------------------------------------------

\section{Related works}

\paragraph{Backpropagating optimization problems}

%As alluded to above, there are several works that incorporate optimizer blocks in deep neural network architectures, and perform differentiation of the optimization routines for backpropagation learning. A subset of these methods applies implicit differentiation~\cite{gould2016differentiating,amos2017optnet}, which we will describe in detail later. Another group of methods conduct numerical differentiation by central differencing: DSAC~\cite{Brachmann2017dsac}, ESAC~\cite{brachmann2019expert}, and NG-DSAC~\cite{brachmann2019neural}. In fact, since these works aim to solve camera localization, they also incorporate a PnP solver in their pipeline. \hl{Something intelligent, convincing and in (a bit more) detail about central differencing and why it is bad relative to implicit differentiation.}

As alluded to above, there are several works that incorporate optimizer blocks in deep neural network architectures, and perform differentiation of the optimization routines for backpropagation learning. A subset of these works address the chellange of incorporating RANSAC in an end-to-end trainable pipeline, such as DSAC~\cite{Brachmann2017dsac}, ESAC~\cite{brachmann2019expert}, and NG-DSAC~\cite{brachmann2019neural}. In fact, since these works aim to solve camera localization, they also incorporate a PnP solver in their pipeline. To backpropagate through the PnP solver, they use central differences to compute the partial derivatives. In effect, if the input dimension is $n$, % and the output dimension is $m$, 
it requires solving PnP $2n$ times in order to obtain the full Jacobian. Another group of methods applies implicit differentiation~\cite{gould2016differentiating,amos2017optnet}, which provides an exact and efficient solution for backpropagating through an optimization process. We will describe implicit differentiation in detail later.

\paragraph{Pose estimation from images}

A target application of our BPnP is pose estimation. Existing works on end-to-end pose estimation~\cite{kendall2015posenet,Kendall2016modelling,kendall2017geometric,brahmbhatt2018geometry,naseer2017deep,walch2017image,wu2017delving} usually employ fully connected layers to compute the target output (pose) using feature maps from previous layers. The output loss function is typically defined using pose metrics (e.g., chordal distance), which are backpropagated using standard differentiation. A recent analysis~\cite{Sattler2019understanding} suggests that what is being performed by these end-to-end networks is akin to learning a set of base poses from the training images, computing a set of weights for the testing image, then predicting the pose as a weighted combination of the base poses. It was further shown that such methods were more related to image retrieval than intrinsically learning to predict pose, hence they may not outperform an image retrieval baseline~\cite{Sattler2019understanding}.

Other pose estimation approaches that combine deep learning with geometric optimization (PnP solver)~\cite{pavlakos20176, rad2017bb8, tekin2018real, Peng2019pvnet, Chen2019satellite} adopt a two-stage strategy: first learn to predict the 2D landmarks or fiducial points from the input image, then perform pose estimation by solving PnP on the 2D-3D correspondences. While the first stage can  benefit from the regularities existing in a training dataset, the second stage (PnP solving) which encodes the fundamental geometric properties of the problem do not influence the learning in the first stage. Contrast this to our BPnP which seamlessly connects both stages, and allows the PnP optimizer to guide the weight updates in the first stage (in addition to standard keypoint or landmark regression losses).

\paragraph{Depth estimation and 3D reconstruction} There exist many works that employ deep networks to learn to predict depth or 3D structure from input images in an end-to-end fashion. Some of these works~\cite{Handa2016gvnn, Clark2018learning, Kendall2017end} can only impose constraints on pairs of images, while others~\cite{Zhou2017unsupervised, Ummenhofer2017demon} learn the structure and the motion in different network branches and do not impose explicit geometric constraints. Also, many of such works~\cite{Eigen2014depth, Ladicky2014pulling, Liu2016learning, Liu2015deep, Laina2016deeper, Xu2017multi, Wang2015towards} require training datasets with ground truth depth labels, which can be expensive to obtain. The proposed BPnP may help to alleviate this shortcoming; as we will show in Sec.~\ref{sec:sfm}, a simple structure-from-motion (SfM) framework that utilizes BPnP can jointly optimize using multiple views (not just two), explicitly impose geometric constraints, and learn structure and motion in an unsupervised fashion without depth labels or ground truth 3D structures.

%\hl{BA-net}~\cite{tang2018ba} - What to say here?

%------------------------------------------------------------------------

\section{Backpropagating a PnP solver (BPnP)}\label{sec:bpnp}

Let $g$ denote a PnP solver in the form of a ``function"
\begin{equation}\label{eq:g}
\bm y = g(\bm x, \bm z, \bK),
\end{equation}
which returns the 6DOF pose $\by$ of a camera with intrinsic matrix $\bK \in \mathbb{R}^{3 \times 3}$ from $n$ 2D-3D correspondences
\begin{align}\label{eq:points}
\bx &= \left[ \begin{matrix} \bx_1^T & \bx_2^T & \dots & \bx_n^T \end{matrix} \right]^T \in \mathbb{R}^{2n \times 1},\\
\bz &= \left[ \begin{matrix} \bz_1^T & \bz_2^T & \dots & \bz_n^T \end{matrix} \right]^T \in \mathbb{R}^{3n \times 1},
\end{align}
where $(\bx_i,\bz_i)$ is the $i$-th correspondence. Let $\pi(\cdot|\bm y, \bK)$ be a projective transformation of 3D points onto the image plane with pose $\bm y$ and camera intrinsics $\bK$. Intrinsically, the ``evaluation" of $g$ requires solving the optimization problem
\begin{align}\label{eq:pnp0}
\begin{aligned}
\by = \argmin_{\by \in SE(3)} \quad & \sum^{n}_{i=1} \left\| \bm r_i \right\|_2^2,
\end{aligned}
\end{align}
where 
\begin{align}
\bm r_i = \bx_i - \bm \pi_i
\end{align}
is the reprojection error of the $i$-th correspondence and
\begin{equation}
    \bm \pi_i = \pi(\bz_i|\by,\bK)
\end{equation}
is the projection of 3D point $\bz_i$ on the image plane. We introduce the shorthand
\begin{align}
\bm \pi := \left[  \bm \pi_1^T, ..., \bm \pi_n^T  \right]^T,
\end{align}
thus~\eqref{eq:pnp0} can be rewritten as
\begin{align}\label{eq:pnp}
\begin{aligned}
\by = \argmin_{\by \in SE(3)} \quad & \left\| \bx - \bm \pi \right\|_2^2.
\end{aligned}
\end{align}
The choice of formulation~\eqref{eq:pnp} will be justified in Sec.~\ref{sec:formulation}.

Our ultimate goal is to incorporate $g$ in a learnable model, where $\bx$, $\bz$ and $\bK$ can be the (intermediate) outputs of a deep network. Moreover, the solver for~\eqref{eq:pnp} should be used to participate in the learning of the network parameters. To this end, we need to treat $g$ as if it were a differentiable function, such that its ``gradients" can be backpropagated to the rest of the network. In this section, we show how this can be achieved via implicit differentiation.

\subsection{The Implicit Function Theorem (IFT)}
 
\begin{tm}[\cite{Krantz2012implicit}]\label{tm:ift}
Let $f:\mathbb R^{n+m} \rightarrow \mathbb R^m$ be a continuously differentiable function with input $(\bm a, \bm b)\in \mathbb R^n \times \mathbb R^m$. If a point $(\bm a^*, \bm b^*)$ satisfies
\begin{equation}\label{eq:f_con}
     f(\bm a^*, \bm b^*) = \bm 0
\end{equation}
and the Jacobian matrix $\frac{\partial f}{\partial \bm b}(\bm a^*, \bm b^*)$ is invertible, then there exists an open set $U \subset \mathbb R^n$ such that $\bm a^* \in U$ and an unique continuously differentiable function $g(\bm a): \mathbb R^n \rightarrow \mathbb R^m$ such that $\bm b^* = g(\bm a^*)$ and 
\begin{equation}
    f(\bm a', g(\bm a')) = \bm 0 \,, \forall \bm a' \in U \,.
\end{equation}

Moreover, for all $\bm a' \in U$, the Jacobian matrix $\frac{\partial g}{\partial \bm a}(\bm a')$ is given by
  \begin{align}\label{eq:implicit_grad}
      \frac{\partial g}{\partial \bm a}(\bm a') = -\left[\frac{\partial f}{\partial \bm b}(\bm a', g(\bm a'))\right]^{-1} \left[\frac{\partial f}{\partial \bm a}(\bm a', g(\bm a')) \right]\,.
  \end{align}
\end{tm}

The IFT allows computing the derivatives of a function $g$ with respect to its input $\bm a$ without an explicit form of the function, but with a function $f$ constraining $\bm a$ and $g(\bm a)$. 

%Eq.~\eqref{eq:f_con} can be relaxed to  
%\begin{equation}\label{eq:f_con2}
%    f(\bm a^*, \bm b^*) = \bm c\,,
%\end{equation}
%where $\bm c$ is a constant vector that does not depend on $\bm a^*$ or $\bm b^*$. This relaxation does not affect the validity of the IFT since we can take $f' = f - \bm c$ which satisfies Eq.~\eqref{eq:f_con} and $\frac{\partial f}{\partial \bm a} = \frac{\partial f'}{\partial \bm a}$, $\frac{\partial f}{\partial \bm b} = \frac{\partial f'}{\partial \bm b}$.

\subsection{Constructing the constraint function $f$}\label{sec:f}

% To invoke IFT for implicit differentiation, we first need to define the constraint function $f(\bm a, \bm b)$ such that Eq.~\eqref{eq:f_con} is upheld. For our problem, we use all four variables $\bm x$, $\bm y$, $\bm z$ and $\bK$ to construct $f$. But we treat $f$ as a two variables function $f(\bm a, \bm b)$, \ie, $\bm a$ takes values in $\{\bm x, \bm z, \bK\}$  - depending on which partial derivative to obtain, and  $\bm b = \bm y$ (the output pose of $g$). 

To invoke the IFT for implicit differentiation, we first need to define the constraint function $f(\bm a, \bm b)$ such that Eq.~\eqref{eq:f_con} is upheld. For our problem, we use all four variables $\bm x$, $\bm y$, $\bm z$ and $\bK$ to construct $f$. But we treat $f$ as a two variables function $f(\bm a, \bm b)$, in which $\bm a$ takes values in $\{\bm x, \bm z, \bK\}$  - depending on which partial derivative to obtain - and  $\bm b = \bm y$ (\ie, the output pose of $g$).

To uphold Eq.~\eqref{eq:f_con}, we exploit the stationary constraint of the optimization process. Denote the objective function of the PnP solver $g$ as
\begin{equation}
    o(\bm x, \bm y, \bm z, \bK) =  \sum_{i=1}^n \lVert \bm r_i \rVert_2^2.
\end{equation}
Since the output pose $\bm y$ of a PnP solver is a local optimum for the objective function, a stationary constraint can be established by taking the first order derivative of the objective with respect to $\bm y$, \ie,
\begin{equation}\label{eq:f0}
    \left. \frac{\partial o(\bm x, \bm y, \bm z, \bK)}{\partial \bm y} \right\rvert_{\bm y = g(\bm x, \bz, \bK)} = \bm 0. 
\end{equation}

Given an output pose from a PnP solver $\bm y = [y_1,...,y_m]^T$, we construct $f$ based on Eq.~\eqref{eq:f0}, which can be written as
\begin{align}
    f(\bm x, \bm y, \bm z, \bK) &= [f_1,...,f_m]^T, 
\end{align}
where for all $j \in \{1,...,m\}$, 
\begin{align}
    f_j &= \frac{\partial o(\bm x, \bm y, \bm z, \bK)}{\partial y_j} \\
    &= 2 \sum_{i=1}^n \langle \bm r_i, \frac{\partial \bm r_i}{\partial y_j}\rangle \\
    &= \sum_{i=1}^n \langle \bm r_i, \bm c_{ij}\rangle 
\end{align}
with 
\begin{equation}\label{eq:cij}
    \bm c_{ij} = -2\frac{\partial \bm \pi_i}{\partial y_j}. 
\end{equation}

\subsection{Forward and backward pass}\label{sec:formulation}
Our PnP formulation~\eqref{eq:pnp} for $g$ essentially performs least squares (LS) estimation, which is not robust towards outliers (egregious errors in $\bx$, $\bz$ and $\bK$). Alternatively, we could apply a more robust objective, such as incorporating an M-estimator~\cite{Zhang1997parameter} or maximizing the number of inliers~\cite{Fischler1981random}. However, our results suggest that LS is actually more appropriate, since its sensitivity to errors in the input measurements encourages the learning to quickly converge to parameters that do not yield outliers in $\bx$, $\bz$ and $\bK$. In contrast, a robust objective would block the error signals of the outliers, causing the learning process to be unstable. 

Given~\eqref{eq:pnp}, the choice of the solver remains. To conduct implicit differentiation, we need not solve~\eqref{eq:pnp} exactly, since~\eqref{eq:f0} is simply the stationary condition of~\eqref{eq:pnp}, which is satisfied by any local minimum. To this end, we apply the Levenberg-Marquardt (LM) algorithm (as implemented in the \texttt{SOLVEPNP\_ITERATIVE} method in OpenCV~\cite{Opencv2000}), which guarantees local convergence. As an iterative algorithm, LM requires initialization $\by^{(0)}$ in solving~\eqref{eq:pnp}. We make explicit this dependence by rewriting~\eqref{eq:g} as
\begin{equation}\label{eq:g2}
\bm y = g(\bm x, \bm z, \bK, \by^{(0)}).
\end{equation}
We obtain the initial pose $\by^{(0)}$ with RANSAC if it is not provided.

In the backward pass, we first construct $f$ as described in Sec.~\ref{sec:f} to then obtain the Jacobians of $g$ with respect to each of its inputs as
%\begin{align}\label{eq:ff}
%   \frac{\partial \bm y}{\partial \bm a} &= -\left[\frac{\partial f}{\partial \bm y}\right]^{-1} \left[\frac{\partial f}{\partial \bm a} \right],
%\end{align}
\begin{align}
   \frac{\partial g}{\partial \bm x} &= -\left[\frac{\partial f}{\partial \bm y}\right]^{-1} \left[\frac{\partial f}{\partial \bm x} \right], \label{eq:ff1}\\
   \frac{\partial g}{\partial \bm z} &= -\left[\frac{\partial f}{\partial \bm y}\right]^{-1} \left[\frac{\partial f}{\partial \bm z} \right], \label{eq:ff2}\\
   \frac{\partial g}{\partial \bK} &= -\left[\frac{\partial f}{\partial \by}\right]^{-1} \left[\frac{\partial f}{\partial \bK} \right]. \label{eq:ff3}
 \end{align}
Given the output gradient $\triangledown \bm y$, BPnP returns the input gradients 
\begin{align}
    \triangledown \bm x &= \left[\frac{\partial g}{\partial \bm x}\right]^T \triangledown \bm y, \\
    \triangledown \bm z &= \left[\frac{\partial g}{\partial \bm z}\right]^T \triangledown \bm y, \\
    \triangledown \bK &= \left[\frac{\partial g}{\partial \bK}\right]^T \triangledown \bm y.
\end{align}

\subsection{Implementation notes}

The number of dimensions of $\bm y$, i.e., $m$, is dependant on the parameterization of $SO(3)$ within the pose. For example, $m = 6$ for the axis-angle representation, $m = 7$ for the quaternion representation, and $m = 12$ for the rotation matrix representation. Experimentally we found the axis-angle representation leads to the best result, possibly since then $m=6$ is equal to the degrees of freedom.

%We find that the axis-angle representation for the rotation part of the pose works best with BPnP. This is possibly \hl{possible?} since this way the pose representation has dimensionality $m = 6$, which equals its degrees of freedom. 

%Eq.~\eqref{eq:implicit_grad} involves the inversion of the Jacobian $\frac{\partial f}{\partial \bm b}$, which can, on occasions, produces large values, making the gradients unstable for training. To address this issue, we normalize the Jacobian $\frac{\partial g}{\partial \bm a}$ with its Frobenius norm, i.e., 
%\begin{equation*}
%    \frac{\partial g}{\partial \bm a} \leftarrow \frac{\partial g}{\partial \bm a} / \lVert \frac{\partial g}{\partial \bm a} \rVert_F.
%\end{equation*}

We compute the partial derivatives in Eqs.~\eqref{eq:cij}, \eqref{eq:ff1}, \eqref{eq:ff2}, and \eqref{eq:ff3} using the Pytorch autograd package~\cite{Paszke2017automatic}.

% The computation of the partial derivatives in Eq.~\eqref{eq:cij}, \eqref{eq:ff1}, \eqref{eq:ff2} and \eqref{eq:ff3} is done numerically using the Pytorch autograd package~\cite{Paszke2017automatic}. 

%-----------------------------------------------------
\section{End-to-end learning with BPnP}

BPnP enables important geometric vision tasks to be solved using deep networks and PnP optimization in an end-to-end manner. Here, we explore BPnP for pose estimation, SfM and camera calibration, and report encouraging initial results. These results empirically validate the correctness of the Jacobians $\frac{\partial g}{\partial \bm x}$, $\frac{\partial g}{\partial \bm z}$ and $\frac{\partial g}{\partial \bK}$ of the PnP solver $g$ obtained using implicit differentiation in Sec.~\ref{sec:f}.

%We demonstrate \hl{how} end-to-end learning benefits from BPnP as a back-propagatable geometric optimization module on three fundamental \hl{geometric optimization?} problems (pose estimation, keypoint regression, and structure from motion). We believe that this examples can bring inspiration for innovations in end-to-end learning.

This section is intended mainly to be illustrative; in Sec.~\ref{sec:objectpose}, we will develop a state-of-the-art object pose estimation method based on BPnP and also report more comprehensive experiments and benchmarks.

\subsection{Pose estimation}\label{sec:pose_reg}

% Given a function $h$, such as a neural network, with input $I$, such as an image, output $\bm x$, such as 2D keypoint predictions, and trainable parameters $\bm \theta$, we with to supervise the system to learn to output the ground truth pose $\bm y^*$. Algorithm~\ref{alg:pose_reg} describes a baseline algorithm to use the BPnP function $g$ in pose estimation, where $l()$ is the loss function and the 3D structure $\bm z$ and the camera intrinsics $\bK$ are known.

Given a known sparse 3D object structure $\bm z$ and known camera intrinsics $\bK$, a function $h$ (a deep network, e.g., CNN, with trainable parameters $\bm \theta$) maps input image $I$ to a set of 2D image coordinates $\bm x$ corresponding to $\bz$, before $g(\bx,\bz,\bK)$ is invoked to calculate the object pose $\by$. Our goal is to train $h$ to accomplish this task. Since the main purpose of this section is to validate BPnP, it is sufficient to consider a ``fixed input" scenario where there is only one training image $I$ with ground truth pose $\by^*$.

Algorithm~\ref{alg:pose_reg} describes the algorithm for this task. The loss function $l(\cdot)$ has the form
\begin{equation}
    l(\bm x, \bm y) = \left\lVert \pi(\bm z|\bm y, \bK) - \pi(\bm z|\bm y^*, \bK) \right\rVert_2^2 + \lambda R(\bm x, \bm y),
\end{equation}
which is the sum of squared errors between the projection of $\bz$ using the ground truth pose $\by^*$ and current pose $\by$ from PnP (which in turn depends on $\bx$), plus a regularization term
\begin{equation}\label{eq:Rxy}
    R(\bm x, \bm y) = \left\lVert \bm x - \pi(\bm z|\bm y, \bK) \right\rVert_2^2.
\end{equation}
The regularization ensures convergence of the estimated image coordinates $\bx$ to the desired positions (note that the first error component does not impose constraints on $\bx$).

\begin{algorithm}[t]
\caption{Pose estimation.}
\label{alg:pose_reg}
\begin{algorithmic}[1]
\STATE $\by \leftarrow$ Identity pose.
\STATE Randomly initialize $\bm{\theta}$
\WHILE{loss $\ell$ has not converged}
\STATE $\bm x \leftarrow h(I;\bm \theta)$.
\STATE $\bm y \leftarrow g(\bm x, \bm z, \bK, \by)$.
\STATE $\ell \leftarrow l(\bm x, \bm y)$.
\STATE $\bm \theta \leftarrow \bm \theta - \alpha \frac{\partial \ell}{\partial \bm \theta}$. \quad (Backpropagate through PnP)   %\hl{Is this where PnP backprop happens?}
%\STATE $\Tilde{\bm y} \leftarrow \bm y$.
\ENDWHILE
\end{algorithmic}
\end{algorithm}

A main distinguishing feature of Algorithm~\ref{alg:pose_reg} is that one of the gradient flow of $\ell$ is calculated w.r.t.~$\by = g(\bx,\bz,\bK)$ before the gradient of $\by$ is computed w.r.t.~to $\bx$ which is then backpropagated to update $\bm\theta$:
\begin{align}\label{eq:posegrad}
\frac{\partial \ell}{\partial \bm \theta} = \frac{\partial l}{\partial \by} \frac{\partial g}{\partial \bx} \frac{\partial h}{\partial \bm \theta} + \frac{\partial l}{\partial \bx} \frac{\partial h}{\partial \bm \theta}.
\end{align}
The implicit differentiation of $\frac{\partial g}{\partial \bx}$ follows Eq.~\eqref{eq:ff1}.

% \begin{figure}[h]
%     \centering
%     \includegraphics[width = \linewidth]{pose_reg1.pdf}
%     \caption{A toy example of Algorithm~\ref{alg:pose_reg} without the regularization term $R(\bm x, \bm y)$ in the loss. $\bm z$ has 8 structural keypoints and the function $h$ is a modified VGG-11~\cite{Simonyan2015very} network which outputs the 8 2D keypoints $\bm x$. Input of $h$ is a tensor of 1s. (a) is the loss curve. (b) shows the evolution of $\bm y$ presented as $\pi_{\bm y}(\bm z)$. (c) shows the evolution of keypoints $\bm x$. In both (b) and (c) the target locations shown as red square markers are $\pi_{\bm y^*}(\bm z)$. }
%     \label{fig:pose_reg1}
%     \vspace{0.46cm}
%     \includegraphics[width = \linewidth]{pose_reg2.pdf}
%     \caption{Another toy example of Algorithm~\ref{alg:pose_reg} with $R(\bm x, \bm y)$ in the loss. Other setup is the same as Fig.~\ref{fig:pose_reg1}.}
%     \label{fig:pose_reg2}
% \end{figure}

\begin{figure}[t]\centering
    \centering
    \includegraphics[width=\linewidth]{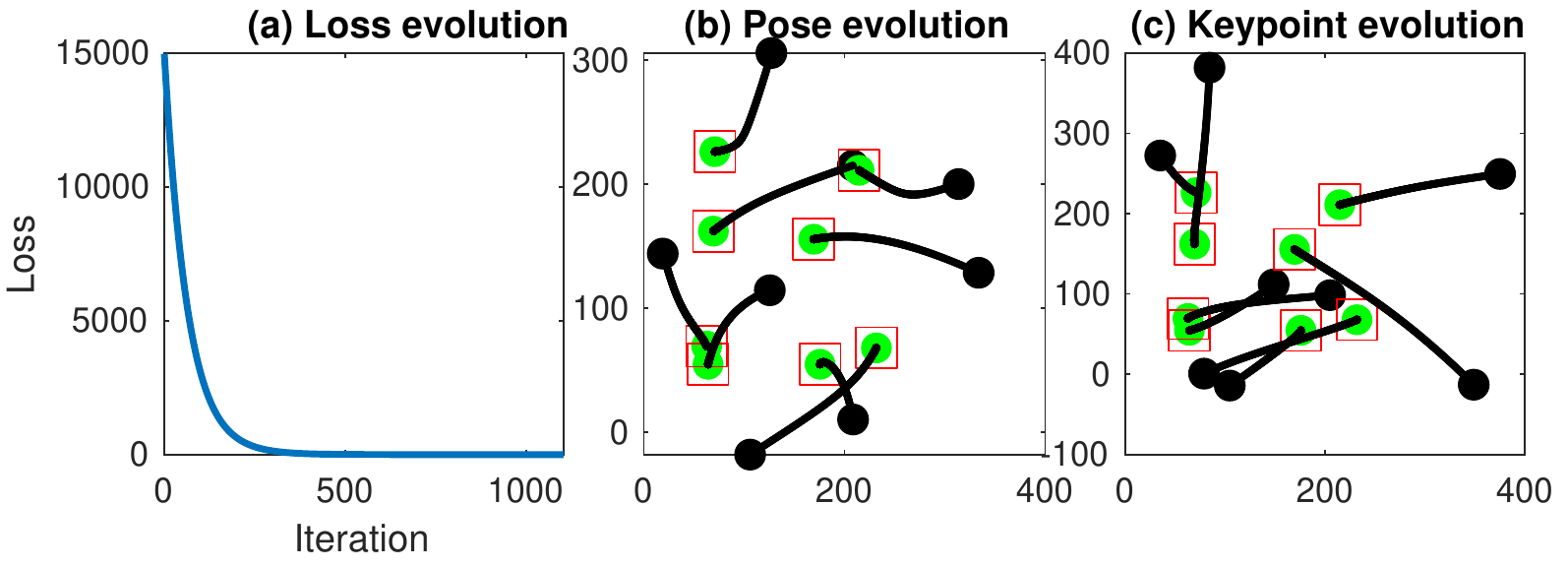}
    \includegraphics[width=\linewidth]{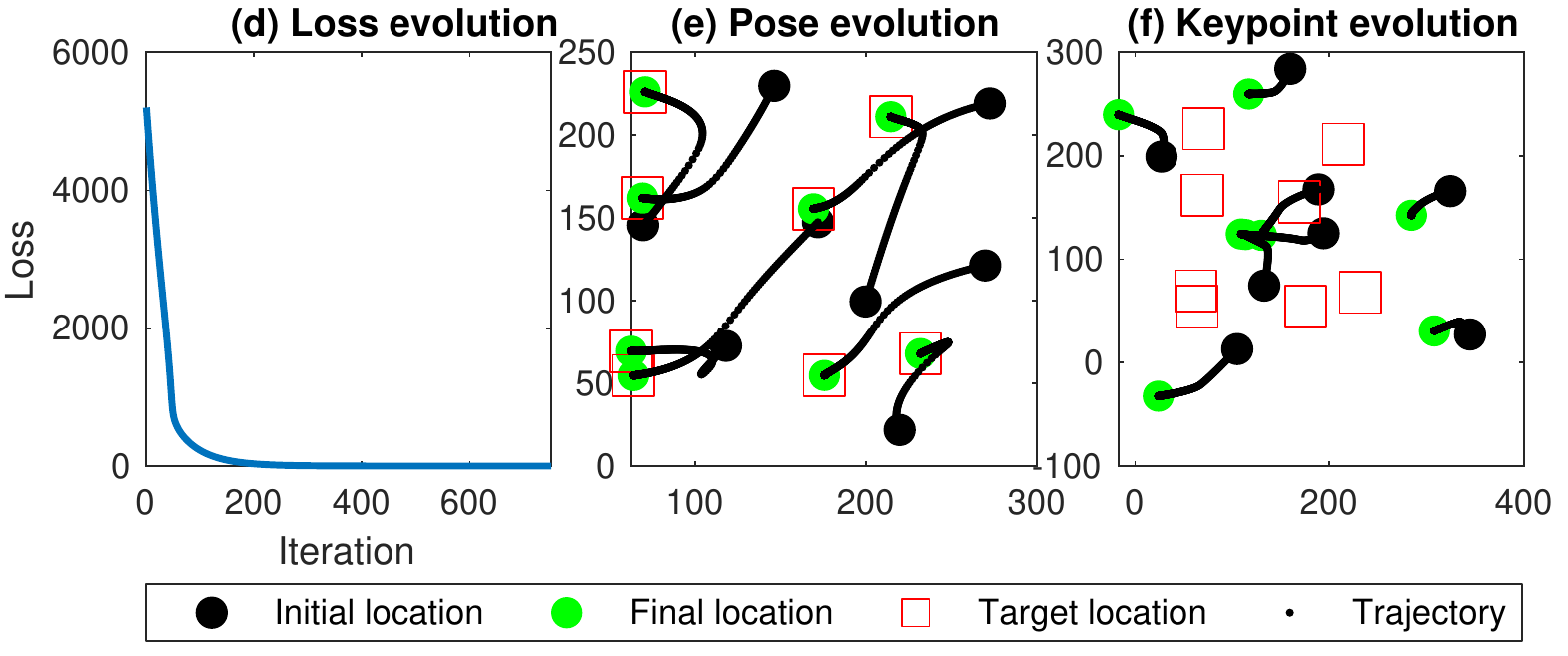}
    \caption{Sample run of Algorithm~\ref{alg:pose_reg} on synthetic data with $n = 8$ landmarks and $h(I;\bm \theta) = \bm \theta$, where $\bm \theta \in \mathbb R^{8\times2}$. The first and second row has $\lambda = 1$ and $\lambda = 0$ respectively. Left column: loss curve. Middle column: evolution of $\bm y$ presented as $\pi(\bm z|\bm y, \bK)$. Right column: the evolution of predicted keypoints $\bm x$. Red square markers represent the target locations $\pi(\bm z|\bm y^*, \bK)$.}
    \label{fig:pose_reg1}
\end{figure}

\begin{figure}[t]\centering
    \centering
    \includegraphics[width=\linewidth]{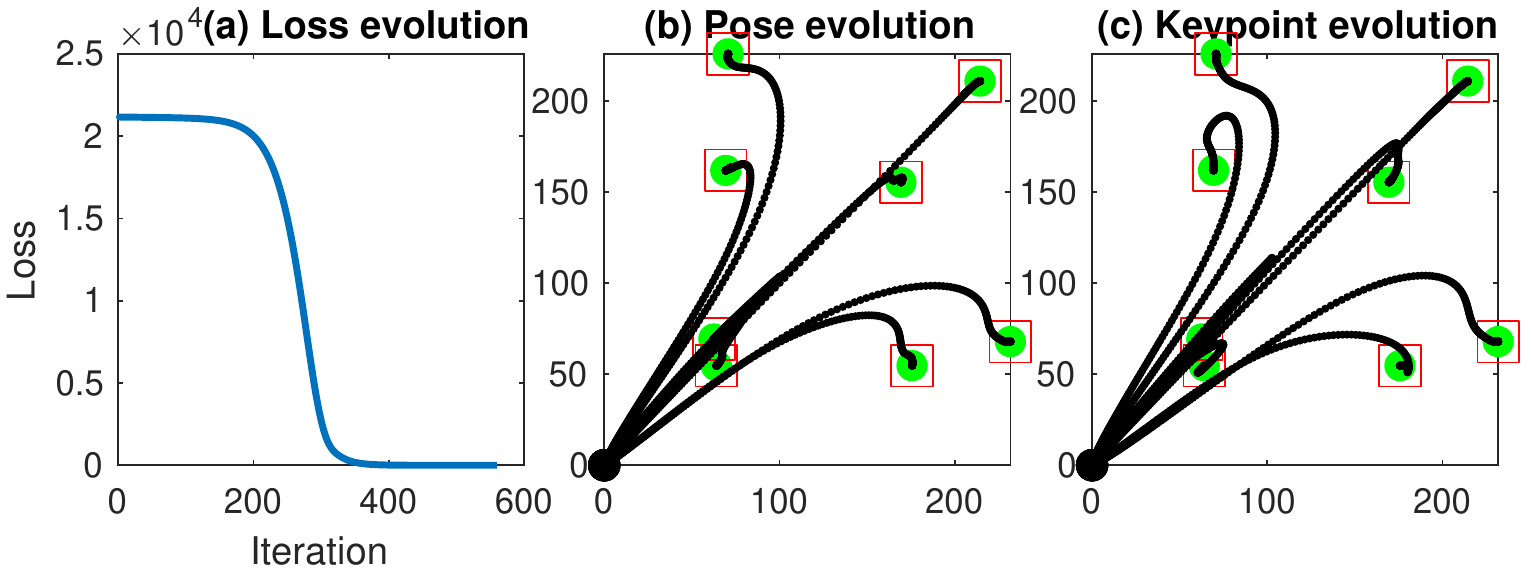}
    \includegraphics[width=\linewidth]{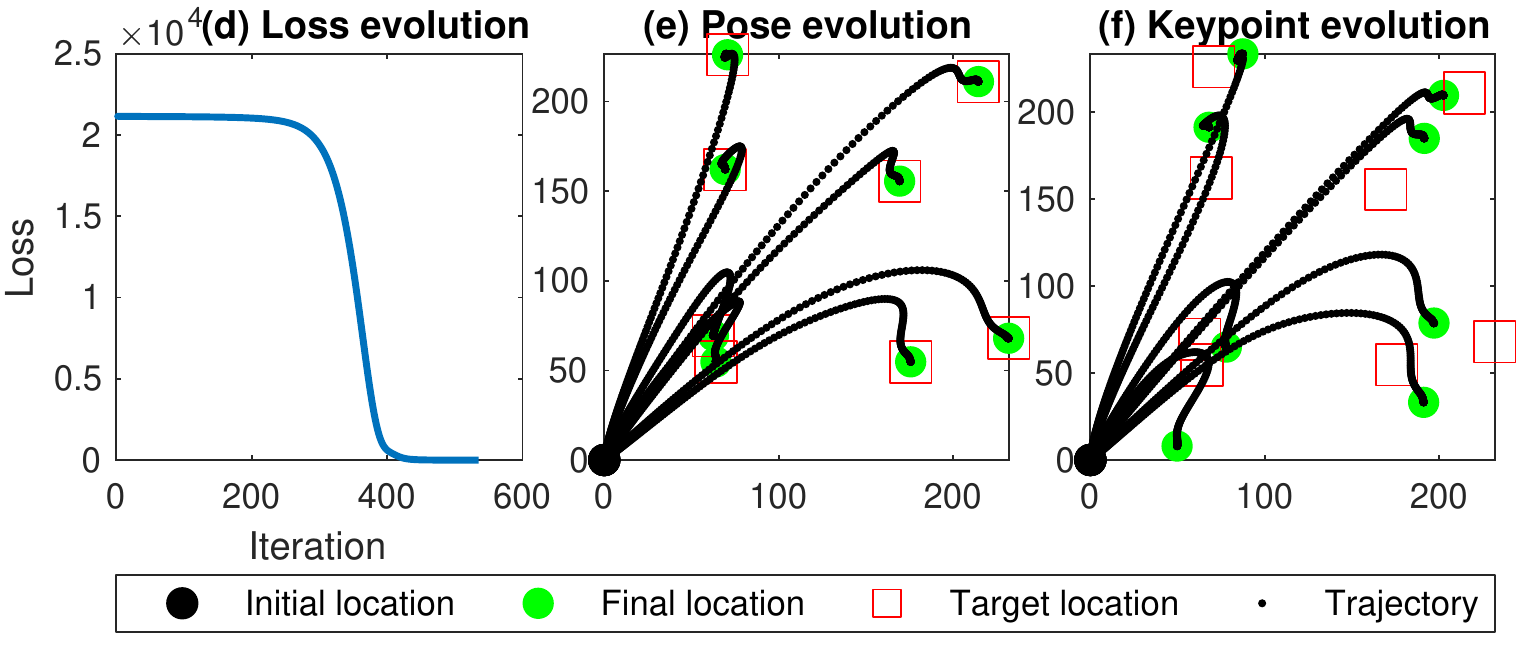}
    \caption{Same experiment as in Fig.~\ref{fig:pose_reg1} except that $h$ is a modified VGG-11~\cite{Simonyan2015very} network which outputs the 2D keypoints $\bm x$.}
    \label{fig:pose_reg2}
\end{figure}

Figs.~\ref{fig:pose_reg1} and \ref{fig:pose_reg2} illustrate Algorithm~\ref{alg:pose_reg} on a synthetic example with $n = 8$ landmarks, respectively for the cases
\begin{itemize}[leftmargin=1em,topsep=0.5em,parsep=0.5em]
\item $h(I;\bm{\theta}) = \bm{\theta}$, i.e., the parameters $\bm{\theta}$ are directly output as the predicted 2D keypoint coordinates $\bx$; and
\item $h(I;\bm{\theta})$ is a modified VGG-11~\cite{Simonyan2015very} network that outputs the 2D keypoints $\bm x$ for $I$.
\end{itemize}
The experiments show that the loss $\ell$ is successfully minimized and the output pose $\by$ converges to the target pose $\by^*$---this is a clear indication of the validity of~\eqref{eq:ff1}.

The experiments also demonstrate the usefulness of the regularization term. While the output pose $\bm y$ will converge to $\bm y^*$ with or without $R(\bm x, \bm y)$, the output of $h$ (the predicted keypoints $\bx$) can converge away from the desired positions $\pi(\bm z|\bm y^*, \bK)$ without regularization.

%\begin{equation}
%    \bm y^* = \argmin_{\bm y} \left\lVert \pi(\bm z|\bm y, \bK) - \bm x \right\rVert_F^2,
%\end{equation}
%that is, any $\bm x$ whose reprojection error happens to be minimized by $\bm y^*$. This will cause the network $h$ to output keypoints inconsistently, affecting convergence in training and generalization in testing. Hence, regularization is needed to ensure that $\bm x$ converges to $\pi(\bm z|\bm y^*, \bK)$.

% Fig.~\ref{fig:pose_reg1} and Fig.~\ref{fig:pose_reg2} demonstrate the effect of the regularization term $R(\bm x, \bm y)$. With or without $R(\bm x, \bm y)$, the output pose $\bm y$ will converge to $\bm y^*$. However, without $R(\bm x, \bm y)$ the output of $h$ can converge to any $\bm x$ whose reprojection error happens to be minimized by $\bm y^*$.

\subsection{SfM with calibrated cameras}\label{sec:sfm}

Let $\{\bm x^{(j)}\}_{j=1}^N$ indicate a set of 2D image features corresponding to $n$ 3D points $\bz$ associated/tracked across $N$ frames $\{ I_j \}^{N}_{j=1}$. Following~\eqref{eq:points}, each $\bx^{(j)}$ is a vector of 2D coordinates; however, $\bz$ may not be fully observed in $I_j$, thus $\bx^{(j)}$ could contain fewer than $n$ 2D coordinates. Let
\begin{equation}
    \bm z^{(j)} = {\cal S}(\bm z | \bm x^{(j)})
\end{equation}
indicate the selection of the 3D points $\bz$ that are seen in $I_j$. Given $\{\bm x^{(j)}\}_{j=1}^N$ and the camera intrinsics for each frame (assumed to be constant $\bK$ without loss of generality), we aim to estimate the 3D structure $\bz \in \mathbb{R}^{3n \times 1}$ and camera poses $\{ \by^{(j)} \}^{N}_{j=1}$ corresponding to the $N$ frames.

\begin{algorithm}[t]
\caption{SfM with calibrated cameras.}
\label{alg:sfm}
\begin{algorithmic}[1]
\STATE $\by^{(j)} \leftarrow$ Identity pose for $j = 1,...,N$.
\STATE Randomly initialize $\bm{\theta}$
\WHILE{loss $\ell$ has not converged}
\STATE $\bz \leftarrow h(\mathbf{1}^{\otimes}; \bm\theta)$.
\STATE $\bm z^{(j)} \leftarrow {\cal S}(\bm z | \bm x^{(j)})$, for $j=1,\dots,N$
\STATE $\bm y^{(j)} \leftarrow g(\bm x^{(j)}, \bm z^{(j)}, \bK, \by^{(j)})$, for $j=1,\dots,N$.
\STATE $\ell \leftarrow l(\{ \by^{(j)} \}_{j=1}^N,\bz)$.
\STATE $\bm \theta \leftarrow \bm \theta - \alpha \frac{\partial \ell}{\partial \bm \theta}$. \quad (Backpropagate through PnP) 
%\STATE $\Tilde{\bm y}^{(i)} \leftarrow \bm y^{(i)},$\, for $i = 1,...,N$
\ENDWHILE
\end{algorithmic}
\end{algorithm}

Our end-to-end method estimates the 3D structure
\begin{align}\label{eq:vggtensor}
\bz = h(\mathbf{1}^{\otimes} ; \bm\theta)
\end{align}
using a deep network $h$ (a modified VGG-11~\cite{Simonyan2015very}) with the input fixed to a $1$-tensor (more on this below); see Algorithm~\ref{alg:sfm}. Note that the algorithm makes use of the PnP subroutine to estimate each camera pose given the current $\bz$ estimate. The loss function $l(\cdot)$ has the form
\begin{align}
l(\{ \by^{(j)} \}_{j=1}^N,\bz) = \sum_{j=1}^N \rVert \bm x^{(j)} - \pi(\bm z^{(j)}|\bm y^{(j)}, \bK) \rVert_2^2,
\end{align}
which is simply the sum of squared reprojection errors across all frames. Again, a unique feature of our pipeline is the backpropagation of the loss through the PnP solver to update network parameters $\bm\theta$.
\begin{align}
    \frac{\partial \ell}{\partial \bm \theta} &= \sum_{j=1}^N \left(\frac{\partial l}{\partial \bz^{(j)}} \frac{\partial \bz^{(j)}}{\partial \bm \theta} + \frac{\partial l}{\partial \by^{(j)}} \frac{\partial \by^{(j)}}{\partial \bz^{(j)}} \frac{\partial \bz^{(j)}}{\partial \bm \theta} \right)
\end{align}
The implicit differentiation of $\frac{\partial \by^{(j)}}{\partial \bz^{(j)}}$ follows Eq.~\eqref{eq:ff2}.

%A significant impact of BPnP is allowing conducting structure from motion in an end-to-end fashion, as simply as shown in Algorithm~\ref{alg:sfm}. Given a track of 2D keypoints from $N$ different viewpoints $\{\bm x^{(i)}\}_{i=1}^N$, we can train a network $h$ to output the 3D structure $\bm z$ by minimizing the reprojection error loss. 

% One significant impact of BPnP is that structure from motion can now be done in a end-to-end fashion, as simply as shown in Algorithm~\ref{alg:sfm}. Given a track of 2D keypoints in $N$ images $\{\bm x^{(i)}\}_{i=1}^N$, we can train a network $h$ to output the 3D structure $\bm z$ by minimizing the projection error loss. 

% In Algorithm~\ref{alg:sfm}, if the initial pose $\bm y_0$ is not provided (\ie, let $\bm y_0 \leftarrow $ None), BPnP will invoke RANSAC to obtain the initial pose for the LM optimization. We found that a good strategy to reduce the chance of converging to a local minimum is to let $\bm y_0 = [0,0,0,0,0,c]$, where $c$ is a constant that is well larger than the ground truth, such as 999. 

\begin{figure}[t]
    \centering
    \includegraphics[width = \linewidth]{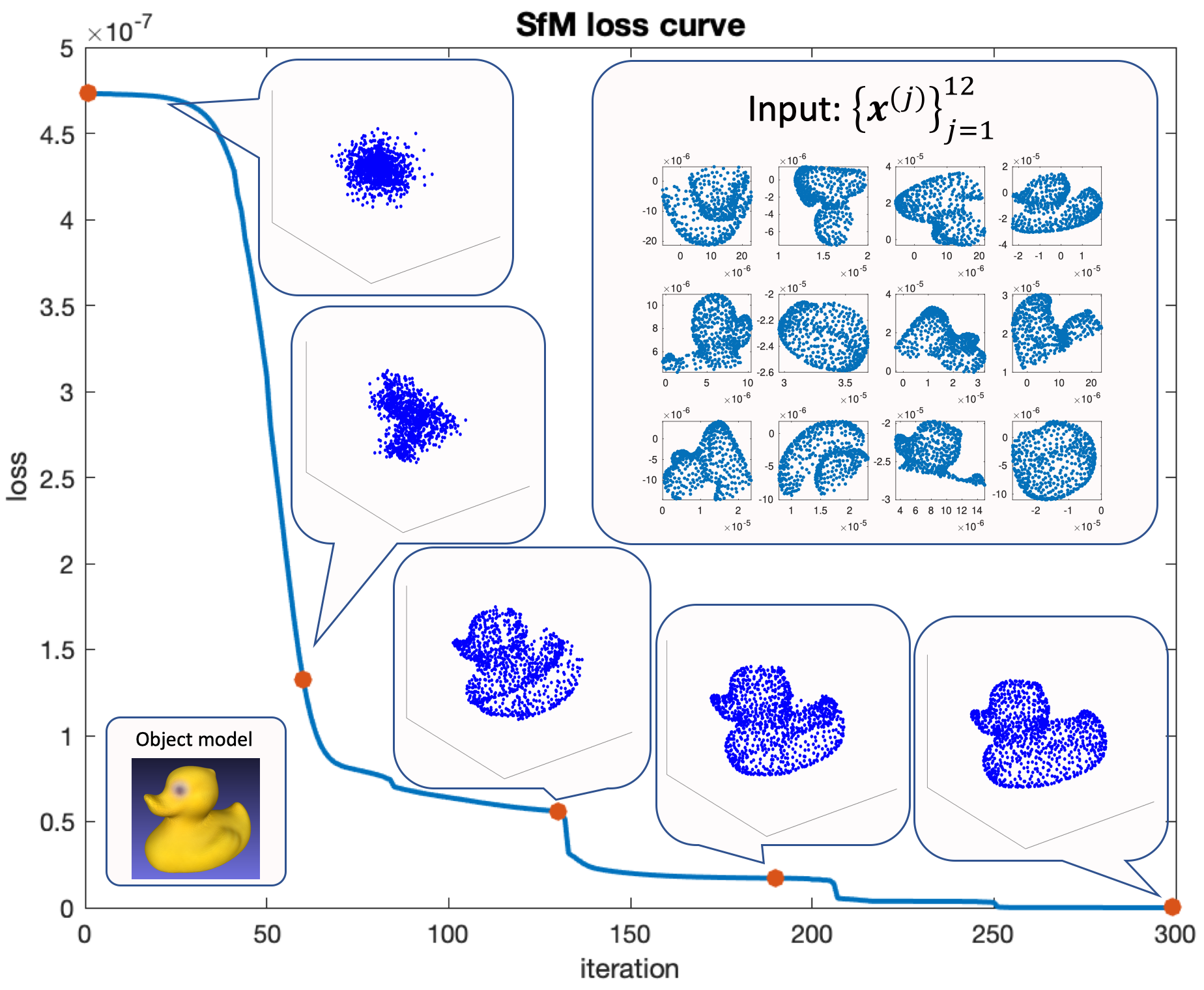}
    \caption{SfM result with Algorithm~\ref{alg:sfm}. The mesh of the object has $n = 1000$ points $\bz^*$, which were projected to $N=12$ different views to obtain $\{\bm x^{(j)}\}_{j=1}^N$ (about half of the 3D points are seen in each view). The function $h$ is a modified VGG-11 network~\cite{Simonyan2015very} which outputs the 3D structure $\bm z$ from a fixed input of $1$-tensor. We depict the output structure $\bm z$ at various steps. A movie of this reconstruction is provided in the supplementary material.}
    \label{fig:sfm_duck}
\end{figure}

\begin{figure}%[t]
    \centering
    \includegraphics[width = \linewidth]{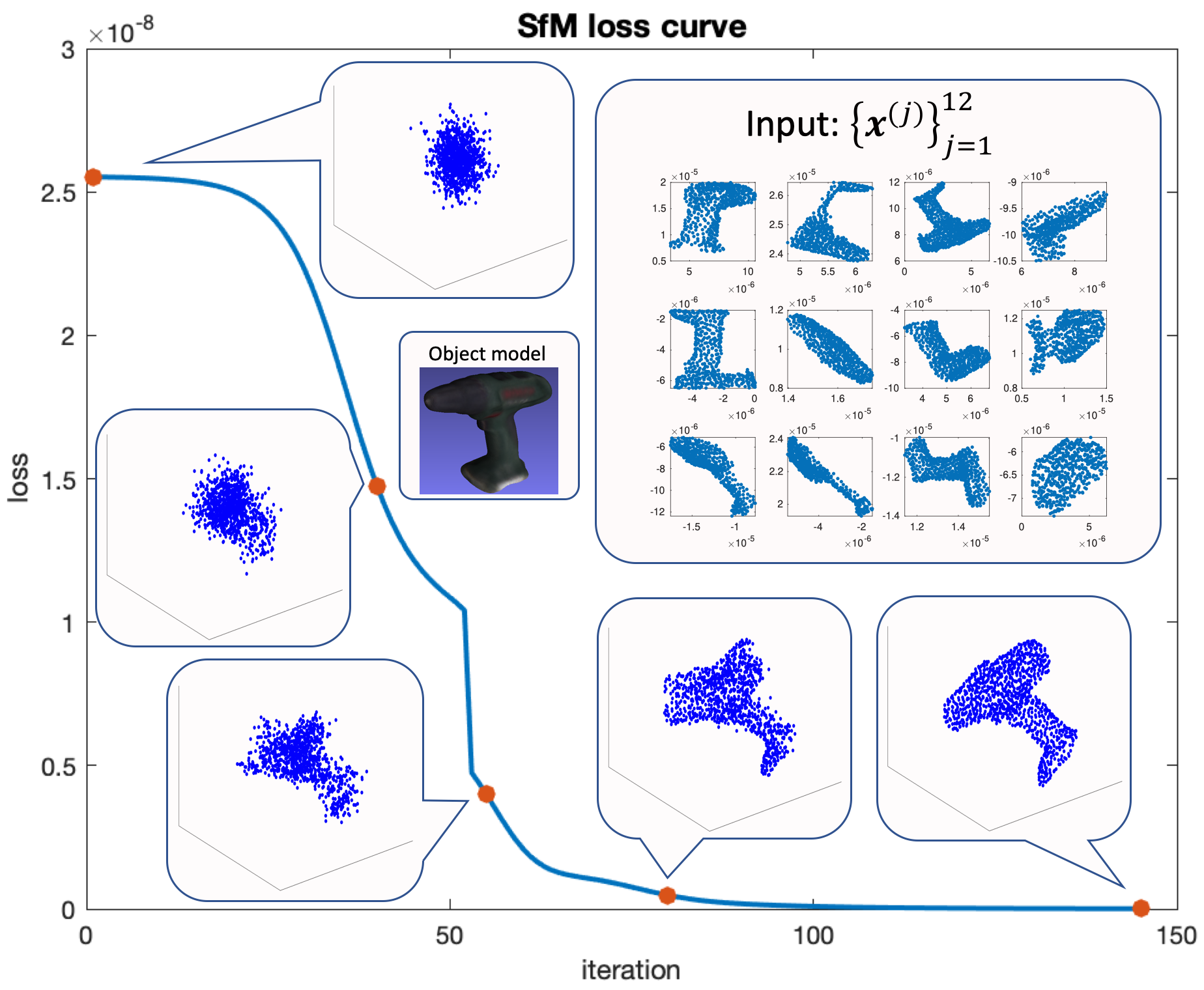}
    \caption{SfM result with a different object which has $n = 1000$ points $\bz^*$. All settings are the same as in Fig.~\ref{fig:sfm_duck}. A movie of this reconstruction is provided in the supplementary material.}
    \label{fig:sfm_driller}
\end{figure}

Fig.~\ref{fig:sfm_duck} illustrates the results of Algorithm~\ref{alg:sfm} on a synthetic dataset with $n = 1000$ points on a 3D object seen in $N = 12$ images (about half of the 3D points are seen in each image). Starting from a random initialization of $\bm\theta$ (which leads to a poor initial $\bz$), the method is able to successfully reduce the loss and recover the 3D structure and camera poses. Fig~\ref{fig:sfm_driller} shows the result from another dataset.

Effectively, our tests show that a generic deep model (VGG-11 with fixed input~\eqref{eq:vggtensor}) is able to ``encode" the 3D structure $\bz$ of the object in the network weights, even though the network is not designed using principles from multiple view geometry. Again, our aim in this section is mainly illustrative, and Algorithm~\ref{alg:sfm} is not intended to replace established SfM algorithms, e.g.,~\cite{schoenberger2016pixelwise, schoenberger2016structure}. However, the results again indicate the correctness of the steps in Sec.~\ref{sec:bpnp}.

\subsection{Camera calibration}

In the previous examples, the intrinsic matrix $\bK$ is assumed known and only $\bx$ and/or $\bz$ are estimated. Here in our final example, given $\bm x$ and $\bm z$ (2D-3D correspondences), our aim is to estimate $\bK$ of the form
\begin{align}
\bK = \left[ \begin{matrix} f_x & 0 & c_x \\ 0 & f_y & c_y \\ 0 & 0 & 1 \end{matrix} \right],
\end{align}
where $f_x$ and $f_y$ define the focal length, and $c_x$ and $c_y$ locate the principal point of the image.

\begin{algorithm}[t]
\centering
\caption{Camera calibration.}
\label{alg:CamCali}
\begin{algorithmic}[1]
\STATE $\by \leftarrow$ Identity pose.
\STATE Randomly initialize $\bm\theta$
\WHILE{loss $\ell$ has not converged}
\STATE $[f_x, f_y, c_x, c_y]^T \leftarrow h(\bm \theta)$.
\STATE $\bK \leftarrow \left[[f_x,0,0]^T~[0,f_y,0]^T~[c_x,c_y,1]^T\right]$.
\STATE $\bm y \leftarrow g(\bm x, \bm z, \bK, \by)$.
\STATE $\ell \leftarrow l(\bK, \by)$.
\STATE $\bm \theta \leftarrow \bm \theta - \alpha \frac{\partial \ell}{\partial \bm \theta}$. \quad (Backpropagate through PnP) 
\ENDWHILE
\end{algorithmic}
\end{algorithm}

We assume $[f_x, f_y, c_x, c_y]^T \in [0,1000]^4$. Under our BPnP approach, we train a simple neural network
\begin{align}
[f_x, f_y, c_x, c_y]^T = h(\bm \theta) = 1000 \,\text{sigmoid}(\bm \theta)
\end{align}
to learn the parameters from correspondences $\bx$ and $\bz$, where $\bm \theta \in \mathbb R^4$. Algorithm~\ref{alg:CamCali} summarizes a BPnP approach to learn the parameters $\bm\theta$ of $h$. The loss function is simply the sum of squared reprojection errors
\begin{align}
l(\bK, \by) = \rVert \bm x - \pi(\bm z|\bm y, \bK) \rVert_2^2,
\end{align}
which is backpropagated through the PnP solver via
\begin{equation}
    \frac{\partial \ell}{\partial \bm \theta} = \frac{\partial l}{\partial \bK} \frac{\partial \bK}{\partial \bm \theta} + \frac{\partial l}{\partial \by} \frac{\partial g}{\partial \bK}\frac{\partial \bK}{\partial \bm \theta}.
\end{equation}
The implicit differentiation of $\frac{\partial g}{\partial \bK}$ follows Eq.~\eqref{eq:ff3}.
%The converged $\bm\theta$ is simply taken as the optimized intrinsic parameters.

\begin{figure}[t]
    \centering
    \includegraphics[width=\linewidth]{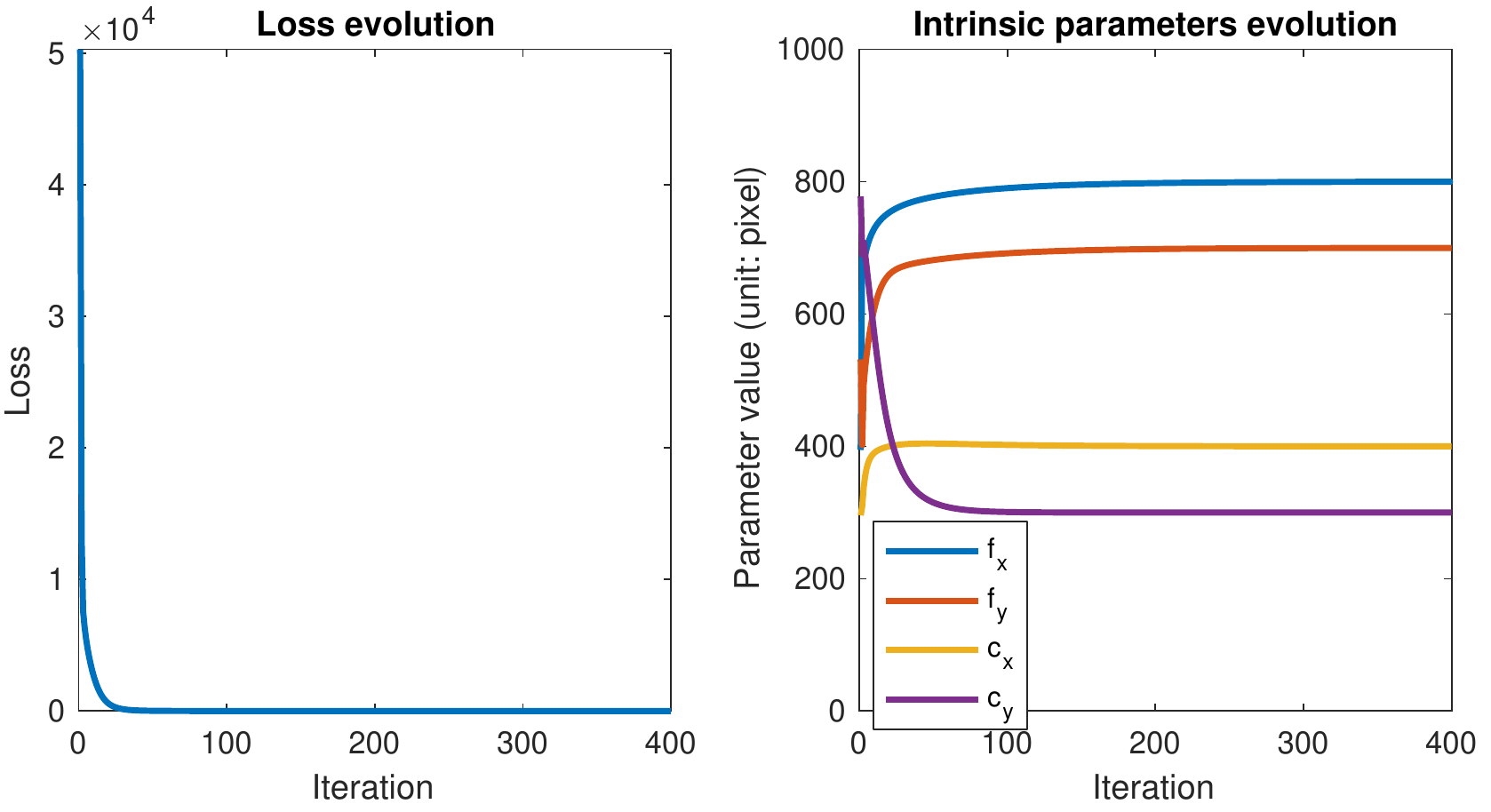}
    \caption{Camera calibration using Algorithm~\ref{alg:CamCali}, based on the ground truth correspondences $\bx$, $\bz$ in Fig.~\ref{fig:pose_reg1}. Left: loss curve. Right: the intrinsic parameters which converge to the ground truth. }
    \label{fig:CamCali}
\end{figure}

Fig.~\ref{fig:CamCali} illustrates the result of Algorithm~\ref{alg:CamCali} using the ground truth correspondences $\bx$, $\bz$ in Fig.~\ref{fig:pose_reg1} as input. The correct intrinsic parameters are $f_x^* = 800, f_y^* = 700, c_x^* = 400, c_y^* = 300$, which the algorithm can clearly achieve as the loss converges to $0$.

%Given the ground truth 2D-3D correspondences $\bm x^*$ and $\bm z^*$, one can utilize the partial derivatives of $\bm y$ with respect to $\bK$ to perform camera calibration. Algorithm~\ref{alg:CamCali} describes a way to let a network $h$ learn the camera intrinsic parameters.

%-----------------------------------------------------

\section{Object pose estimation with BPnP}\label{sec:objectpose}

%We apply BPnP in a real-world application by proposing the object pose estimation pipeline depicted in Fig.~\ref{fig:obpose_pipeline}, which directly regresses the pose from an input image. We use HRNet~\cite{Sun2019deep} as the backbone to predict the landmark heatmaps $\Phi$. We then use the Differentiable Spatial to Numerical Transform (DSNT)~\cite{Nibali2018numerical} to convert heatmaps $\Phi$ to 2D landmark coordinates $\bm x$. Finally, BPnP obtains the pose $\bm y$ from the 2D landmarks $\bm x$ and the 3D structural landmarks of the object $\bm z$.

We apply BPnP to build a model that regresses the object pose directly from an input image, not through fully connected layers, but through projective geometric optimization while remaining end-to-end trainable. Our model is unique in that it simultaneously learns from feature-based loss and geometric constraints in a seamless pipeline. 

\begin{figure} %[h] 
    \centering
    \includegraphics[width=\linewidth]{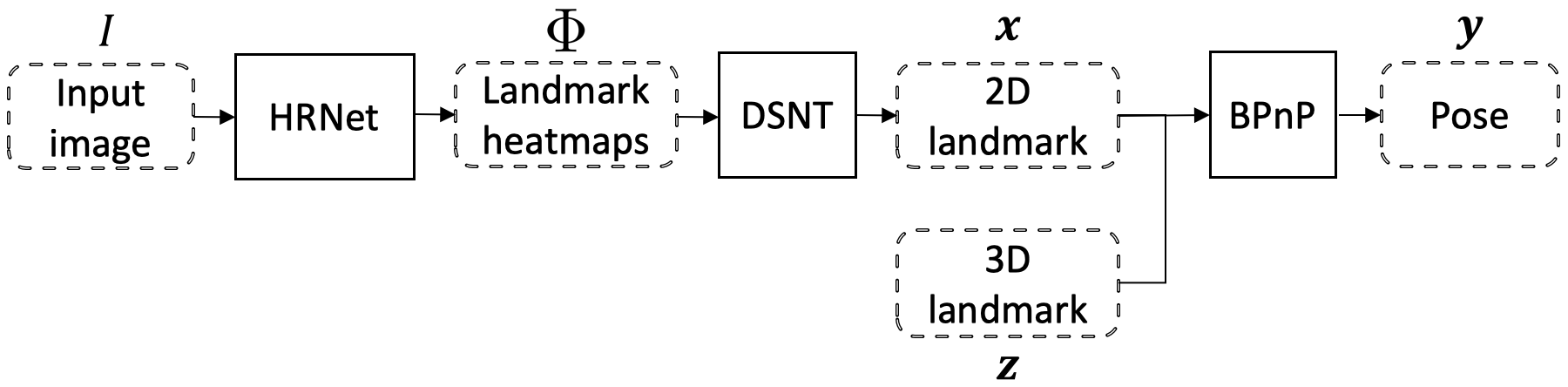}
    \caption{The pipeline of our object pose estimation network. }
    \label{fig:obpose_pipeline}
\end{figure}

The pipeline of the proposed model is depicted in Fig.~\ref{fig:obpose_pipeline}. We use HRNet~\cite{Sun2019deep} as the backbone to predict the landmark heatmaps $\Phi$. We then use the Differentiable Spatial to Numerical Transform (DSNT)~\cite{Nibali2018numerical} to convert heatmaps $\Phi$ to 2D landmark coordinates $\bm x$. Finally, BPnP obtains the pose $\bm y$ from the 2D landmarks $\bm x$ and the 3D structural landmarks of the object $\bm z$.

Let $\Phi^*$ denote the ground truth heatmaps constructed with the ground truth 2D landmarks $\bm x^*$. We define a heatmap loss
\begin{equation}
    \ell_h = \text{MSE}(\Phi,  \Phi^*),
\end{equation}
where MSE$(\cdot,\cdot)$ is the mean squared error function; a pose loss
\begin{equation}
    \ell_p = \left\lVert \pi(\bm z\,|\;\bm y, \bK) - \bm x^* \right\rVert_F^2 + R(\bm x, \bm y);
\end{equation}
and a mixture loss
\begin{equation}
    \ell_m = \ell_h + \beta \left\lVert \pi(\bm z\,|\;\bm y, \bK) - \bm x^* \right\rVert_F^2,
\end{equation}
for training the model respectively. The regularization term $R(\bm x, \bm y)$ is defined in Eq.~\eqref{eq:Rxy}. Note that in the mixture loss $R(\bm x, \bm y)$ is unnecessary because the heatmap loss $\ell_h$ acts as a regularization term. We set the balancing weight $\beta$ to $0.0002$ in the experiments.

We apply our pipeline on the LINEMOD~\cite{Hinterstoisser2012model} dataset. For each object we
\begin{itemize}[leftmargin=1em,topsep=0.3em,parsep=0.3em]
    \item obtain a 3D model representation consisting of 15 landmarks by using the Farthest Point Sampling (FPS)~\cite{Peng2019pvnet} over the original object mesh, 
    \item randomly reserve $400$ images as the test set and set the remaining (about $800$, depending on the object) as the training set, and
    \item train a model to predict the 6DOF object pose from the input image.
\end{itemize}
We train each model with three different losses ($\ell_h$, $\ell_p$, and $\ell_m$), for 120 epochs each. To assist convergence, when training the model with $\ell_p$ and $\ell_m$, we first train with $\ell_h$ for the first 10 epochs leaving the remaining $110$ epochs to train the target loss.

%, then switch the loss to $\ell_p$ and $\ell_m$, respectively, to train for another 110 epochs.

%We train the model with three different losses $\ell_h$, $\ell_p$, and $\ell_m$, for 120 epochs each. To assist convergence, when training the model with $\ell_p$ and $\ell_m$, we first train with $\ell_h$ for the first 10 epochs, then switch the loss to $\ell_p$ and $\ell_m$, respectively, to train for another 110 epochs.

We evaluate our method with the following two metrics. 
\begin{description}
    \item[Average 3D distance of model points (ADD)~\cite{Hinterstoisser2012model}] This is the percentage of accurately predicted poses in the test set. We consider a predicted pose as accurate if the average distance between the 3D model points expressed in the predicted coordinate system and that expressed in the ground truth coordinate system is less than 10\% of the model diameter. For symmetric objects we use the  ADD-S~\cite{Xiang2018posecnn} metric instead which is based on the closest point distance.

    %It is the percentage of accurately predicted poses in the test set. A predicted pose is considered accurate if the average distance between the 3D models transformed by the predicted pose and the ground truth pose, respectively, is less than 10\% of the model diameter. If an object is considered symmetric, the ADD-S~\cite{Xiang2018posecnn} metric is used, in which the closest point distance is used to compute the ADD.
    
    \item [2D projection~\cite{Brachmann2016Uncertainty}.] Mean distance between 2D keypoints projected with the estimated pose and those projected with ground truth pose. An estimated pose is considered correct if this distance is less than a threshold $\psi$. 
\end{description}

\begin{table*} %[h]
    \begin{center}
    \begin{tabular}{|c|cccc|cccc|ccc|}
   \hline
    \multicolumn{1}{|c|}{\multirow{3}{*}{Model} }
   &\multicolumn{4}{c|}{ADD(-S)}
   &\multicolumn{4}{c|}{2D projection with $\psi=5$}
   &\multicolumn{3}{c|}{2D projection with  $\psi=2$}
   \\
   \cline{2-12}
  &\multicolumn{3}{c|}{Ours}
  &\multicolumn{1}{c|}{\multirow{2}{*}{PVNet}}
  &\multicolumn{3}{c|}{Ours}
  &\multicolumn{1}{c|}{\multirow{2}{*}{PVNet}}
  &\multicolumn{3}{c|}{Ours}
  \\
  \cline{2-4}
  \cline{6-8}
  \cline{10-12}
  & \multicolumn{1}{c}{$\ell_h$} & \multicolumn{1}{c}{$\ell_p$} & \multicolumn{1}{c|}{$\ell_m$} &
  & \multicolumn{1}{c}{$\ell_h$} & \multicolumn{1}{c}{$\ell_p$} & \multicolumn{1}{c|}{$\ell_m$} &
  & \multicolumn{1}{c}{$\ell_h$} & \multicolumn{1}{c}{$\ell_p$} & \multicolumn{1}{c|}{$\ell_m$} 
  \\
  \hline
ape&74.00&56.75&\textbf{74.75}&43.62&\textbf{99.50}&\textbf{99.50}&\textbf{99.50}&99.23 & 90.00 & 86.75 & \textbf{93.75} \\
benchwise&98.50&98.00&99.00&\textbf{99.90}&99.25&98.75&99.25&\textbf{99.81}&\textbf{86.00} & 82.75 & \textbf{86.00} \\
cam&\textbf{96.25}&83.75&\textbf{96.25}&86.86&98.75&98.50&\textbf{99.25}&99.21& \textbf{91.50} & 81.50 & 90.50 \\
can&97.00&94.75&\textbf{98.00}&95.47&99.50&99.50&99.75&\textbf{99.90}& \textbf{93.25} & 89.00 & 92.75 \\
cat&93.00&85.25&\textbf{94.25}&79.34&\textbf{99.50}&\textbf{99.50}&\textbf{99.50}&99.30& \textbf{96.75} & 95.25 & \textbf{96.75} \\
driller&98.50&98.00&\textbf{99.25}&96.43&\textbf{99.00}&98.50&98.50&96.92& 83.75 & 81.50 &\textbf{84.50} 
\\
duck&76.25&49.25&\textbf{78.50}&52.58
&99.00&\textbf{99.25}&99.00&98.02
& 88.50 & 84.00 & \textbf{91.50} 
\\
eggbox&95.75&93.25&96.50&\textbf{99.15}
&\textbf{99.50}&99.25&\textbf{99.50}&99.34
& 92.75 & \textbf{93.25} & 92.50 
\\
glue&87.50&76.25&90.00&\textbf{95.66}
&\textbf{99.50}&\textbf{99.50}&\textbf{99.50}&98.45
& 93.50 & 90.00 & \textbf{94.00} 
\\
holepuncher&89.50&80.25&\textbf{91.50}&81.92
&99.75&99.50&99.75&\textbf{100.00}
& \textbf{92.25} & 90.25 & 91.75 
\\
iron&97.75&96.50&97.75&\textbf{98.88}
&98.50&98.25&98.50&\textbf{99.18}
& 86.75 & 79.75 & \textbf{87.25} 
\\
lamp&\textbf{99.75}&98.50&\textbf{99.75}&99.33
&\textbf{98.75}&98.00&98.50&98.27
& 85.00 & 83.25 & \textbf{86.75} 
\\
phone&95.75&96.00&\textbf{97.00}&92.41
&99.25&99.00&99.25&\textbf{99.42}
& 91.50 & 88.00 & \textbf{92.25} \\
\hline
average&92.27&85.12&\textbf{93.27}&86.27
&\textbf{99.21}&99.00&\textbf{99.21}&99.00
& 90.12 & 86.56 & \textbf{90.79} 
\\
\hline
 \end{tabular}
\end{center}
\caption{
Test accuracy on the LINEMOD dataset in terms of the ADD(-S) metric (columns 2-5) and the 2D projection metric with $\psi = 5$ pixels (columns 6-9) and $\psi=2$ pixels (columns 10-12). Objects eggbox and glue are considered as symmetric objects and the ADD-S metric is used. }
\label{tab:result}
\end{table*}

\begin{figure*}%[t]
    \centering
    \includegraphics[width=0.98\linewidth]{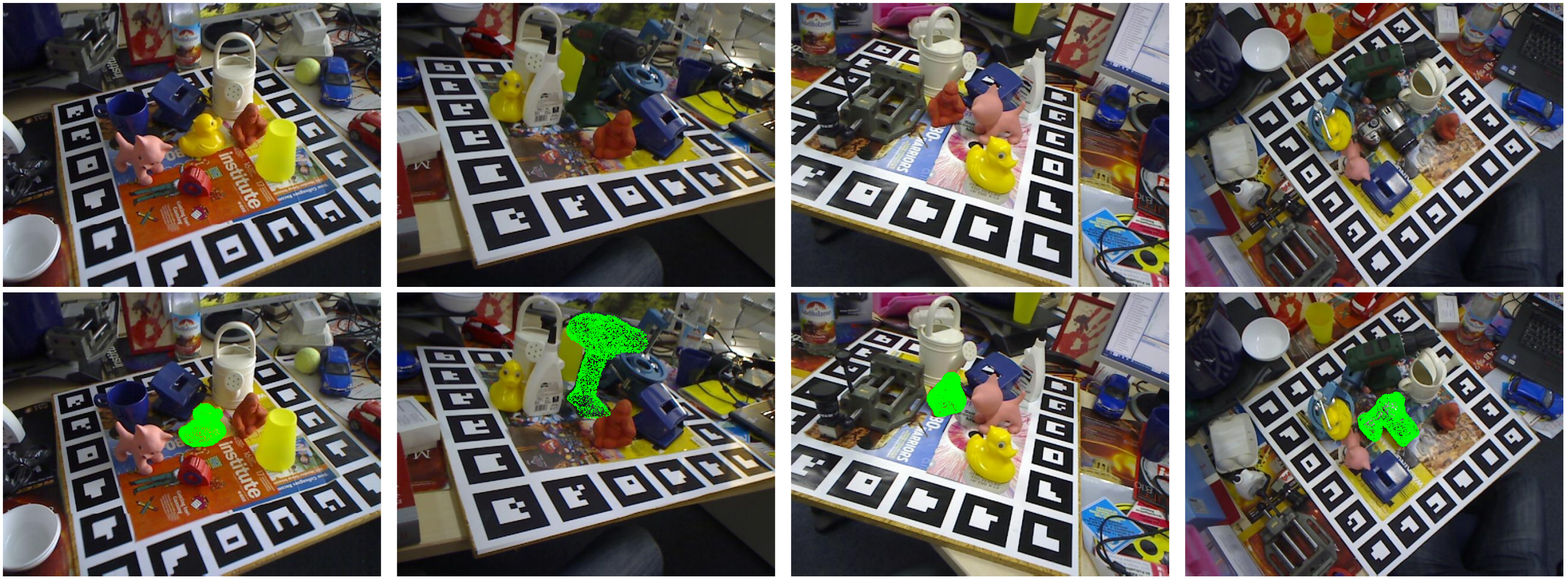}
    \caption{Random sample of test results of the proposed model trained with the mixture loss $\ell_m$. The first row are the test images for predicting the pose of the central object. The second row shows the regressed alignment for each query object. }
    %A random sample of test results of our method trained with mixture loss $\ell_m$. Left column are the test images where the poses of the central objects are to be predicted. Right column shows the projections of the 3D object mesh using the estimated poses.
    \label{fig:linemod}
\end{figure*}

Table~\ref{tab:result} summarizes the results of our experiments. In terms of the ADD(-S) metric, the model trained with $\ell_h$ performs considerably better than the one with $\ell_p$. As expected, heatmaps can exploit richer spatial features than coordinates. However, the mixture loss achieves the highest accuracy, which suggests that heatmap loss benefits from additional correction signals from the pose loss.

%Table~\ref{tab:result} reports the ADD(-S) metric. The model trained with $\ell_h$ performs much better than the one with $\ell_p$. As expected, heatmaps can exploit richer spatial features than coordinates. However, the mixture loss achieves the highest accuracy, which suggests that heatmap loss can benefit from additional correction signals from the pose loss. 

In terms of the 2D projection metric, all methods perform similarly, with an average accuracy of at least 99\%.  To better distinguish the performances amongst different loss functions, we tighten the positive threshold $\psi$ from the standard 5 pixels to 2 pixels. Consistent with the ADD(-S) result,  the mixture loss outperformed pure heatmap loss on training an object pose estimation model. Visualization of a random subset from the test results is shown in Fig.~\ref{fig:linemod}.

% In terms of the 2D projection metric, as shown in Table~\ref{tab:result_2d_5pixel}, all methods perform similarly, achieving an average accuracy of at least 99\%.  To better distinguish the performances amongst different loss functions, we tighten the positive threshold from the standard 5 pixels to 2 pixels, the result of which is reported in Table~\ref{tab:result_2d_2pixel}. Consistent with what we find in Table~\ref{tab:result}, the model trained with the mixture loss $\ell_m$ outperforms the one with pure heatmap loss $\ell_h$.

We provide the result of the current state-of-the-art PVNet~\cite{Peng2019pvnet} as a reference. Overall, models trained with $\ell_h$ and $\ell_m$ have higher average test accuracy than PVNet, in terms of both the ADD(-S) metric and the 2D projection metric. We remind the reader to be aware of several factors while comparing the performances: we use a different backbone from PVNet; our train-test numbers of images are about 800-400 while about 20200-1000 in PVNet. Because we require the ground truth pose label for training, we did not use any data augmentation such as cropping, rotation or affine transformation. 

%We provide the result of the current state-of-the-art PVNet~\cite{Peng2019pvnet} as a reference. Overall, models trained with $\ell_h$ and $\ell_m$ have higher average test accuracy than PVNet. However, we remind the readers to note several different factors: we use a different backbone from PVNet; our train-test number of images are about 800-400 while PVNet's are about 22000-1000. Because we require the ground truth pose label for training, we did not use any data augmentation such as cropping, rotation or affine transformation. 

%----------------------------------------------------

\section{Conclusions}
% In this paper we present BPnP, a novel method to do back-propagation through a PnP solver. We show that the gradients of such geometric optimization process can be computed using the Implicit Function Theorem as if it is differentiable. Furthermore, we develop a residual-conformity trick to make end-to-end pose estimation using BPnP smooth and stable. We also propose a ``march in formation'' algorithm which successfully uses BPnP for keypoint regression. 
  
We present BPnP, a novel approach for performing backpropagation through a PnP solver. BPnP leverages on implicit differentiation to address computing the non-explicit gradient of this geometric optimization process. We validate our approach in three fundamental geometric optimization problems (pose estimation, structure from motion, and camera calibration). Furthermore, we developed an end-to-end trainable object pose estimation pipeline with BPnP, which outperforms the current state-of-the-art. Our experiments show that exploiting 2D-3D geometry constraints improves the performance of a feature-based training scheme.

The proposed BPnP opens a door to vast possibilities for designing new models. We believe the ability to incorporate geometric optimization in end-to-end pipelines will further boost the learning power and promote innovations in various computer vision tasks.

\subsection*{Acknowledgement} This work was supported by ARC LP160100495 and the Australian Institute for Machine Learning. Nan Li is sponsored by NSF China (11601378) and Tencent-SZU Fund.

%Our invention opens a door to vast possibilities on designing new models. We believe the ability of incorporating geometric optimization in end-to-end pipelines will further boost the learning power and promote innovations in various computer vision tasks. 

\clearpage

{\small
\bibliographystyle{ieee_fullname}
\bibliography{egbib}
}

\end{document}